# Planning Method for Skill-Based Control of Robots Using a PLC as Skill Trigger

Andreas Gaugenrieder*[a], Hari Hara Balasubramaniam[a], Jannik Möhrle[a], Rüdiger Daub[a,b]

[a]*Fraunhofer Institute for Casting, Composite and Processing Technology IGCV, Am Technologiezentrum 10, 86159 Augsburg, Germany*
[b]*Institute for Machine Tools and Industrial Management, Technical University of Munich, Bolzmanstraße 15, 85748 Garching b. München, Germany*

* Corresponding author. Tel.: +49-821-90678-313; fax: +49-821-90678-199. E-mail address: andreas.gaugenrieder@igcv.fraunhofer.de

**Abstract**

Skill-based programming of robots provides a flexible approach for automation. Existing solutions neglect the optimization of motion sequences, leading to inefficiencies in execution. This work introduces a planning method that enhances skill-based robot programming by integrating motion sequence optimization. This optimization leads to a new *MoveContinuousSkill*. The software for executing the *MoveContinuousSkill* is implemented on a Programmable Logic Controller and applied across multiple robotic systems. Experimental results demonstrate a significant improvement in execution time through optimized motion sequence.

## 1 Introduction

The increasing demand for High-Mix Low-Volume (HMLV) production necessitates highly flexible modern manufacturing techniques such as Flexible Manufacturing System (FMS) and Modular Production System (MPS) [1]. These systems provide significant hardware flexibility and adaptability. Their full potential is hindered by limitations of their control software, particularly in flexibility, modularity, reusability and scalability [2]. There is an increase in the need for automation due to the general workforce shortage [3]. Additional challenges exist due to the lack of qualified specialists in the fields of information technology and engineering, which means that fewer personnel are available to develop automation solutions [4].

In recent years, skill-based control software has appeared as a promising solution that enables flexible programming and saves time [5]. Skills are solution-neutral functions offered by a device with specific input and output parameters [6]. The integration of skills in the product-process-resource-model leads to time savings in the programming of FMS and MPS [7, 8]. Skill-based control software addresses the workforce challenge by enabling high flexibility, reusability and scalability of software, thereby reducing the dependency on highly qualified specialists [2].

Due to their high degree of kinematic freedom, robots are often used in flexible automation [9]. However, existing approaches to skill-based programming of robots lead to inefficiencies compared to classic motion programming in robot controllers, like longer cycle times, which are discussed in more detail in the chapter 2.

Particularly Programmable Logic Controllers (PLCs) are widely adopted in industry, which reveals that over 91 % of machine builder in the mechanical industries in Germany consider standard PLC programming essential [10]. The wide use of PLC is due to the demands of machine buyer and its technical properties, like real-time and high interface variance [10].

To address the existing inefficiencies in skill-based robot programming, several elements are presented in this work. A *MoveContinuousSkill* is introduced, which can consist of a sequence of one or more motions. Chapter 4 presents a method for using the *MoveContinuousSkill*. To make skill-based programming more popular, it is necessary to apply it to the widely used PLC. Chapter 5 presents an interface and logic that enables flexible control of the robot from the PLC. The paper assesses the performance of the developed skill-based software framework, highlighting its improved performance over the existing approaches in terms of cycle times.

## 2 Related Work

This chapter provides an overview of the robot skills and the methods for robot skill planning used in research. The section on the PLC-Robot Interface shows the solutions used in industry.

### 2.1 Robot Skills

Since robots are essential for advanced manufacturing systems, especially the articulated robots, they are commonly used across various industries. Consequently, extensive research has focused on developing and assigning skills to robots. These skills are highly reusable due to the standardized functionalities and capabilities provided by the robot manufacturers [11]. The earliest approach to robot skills was presented in the scientific work [8], where a generic skill named 'grasp and move with robotic arm' was proposed. This generic assignment of skills of actuator was further refined in the work [12] as '*MoveAbsolute*' that can be assigned to a robot. Another study [13] demonstrated the use of more operation-specific skill '*Grinding*', assigned to a robot equipped with a grinding tool. The significant advancement in the resource-specific robot skills was achieved in [14], which defined skills such as '*CartesianLinearMoveSkillType*', and '*JointPtpMoveSkill-Type*'. These skills were able to perform linear and point-to-point robotic motions based on both cartesian coordinate frame and joint values. Other research works assigned robotic skills in varied ways. For instance, [11] introduced robotic skills such as '*Move to pose*' and '*Move linear*' for moving the Tool Center Point (TCP) of the robot to the specified pose or along a linear path on the desired axis. Meanwhile [15], [16] and [2] described robot motion more abstractly using a generalized '*Move*' skill. In another significant contribution, the study assigned skills of robot as '*MoveAbsolute*', '*MoveRelative*' and '*MoveVelocity*' for robot motion in absolute frame, relative frame and at specified velocity respectively [17]. The robot was triggered by a PLC that was programmed according to IEC 61499 in the Eclipse 4diac framework.

### 2.2 Robot Skill Planning

The flexibility and reusability of skills save significant time and effort by eliminating the need for repetitive programming as these skills can be reused for performing similar functions. Simultaneously, numerous research works were performed to automate the utilizing these skills to increase the efficiency in a system with dynamic requirements [17].

The initial research work for utilizing the skills autonomously based on the process requirement was conducted in [7], which reviewed various process planning frameworks.

While early research primarily focused on overall process planning, [18] placed a stronger emphasis on planning and utilizing robot skills based on CAD model requirements. The functioning of each skill was controlled and monitored by a dedicated state machine, ensuring the execution of the skill when it reaches the final state. Once a skill function is executed, the following skill is triggered by the planning framework. This method of skill monitoring and control through state machine gained prominence in research [11, 14].

Since the state machines are efficient in controlling and monitoring the skills, each skill in this paper and other works had its own dedicated state machine. But this leads to interruptions of motion between each consecutive robot skills, as the subsequent skill is triggered only after the previous one is completed. This interruption, while necessary when the robot tool performs a standstill operation, is undesirable in scenarios where continuous motion is required. Due to the necessary accelerations and decelerations, these interruptions lead to a lower speed when executing a sequence of motions. The interruptions lead to increased cycle times, higher energy consumption and additional wear and tear on robotic transmission systems due to jerks during motion transitions [19].

### 2.3 PLC-Robot Interface

The interfacing and programming of robots into PLC environments commonly relies on vendor-specific interfaces, such as mxAutomation for KUKA robots and MotoLogix for YASKAWA robots. These Profinet-based interfaces allow the programming of robotic functions directly within the PLC environment but significantly limit the flexibility and scalability of skill-based control software. Since these interfaces are specific to manufacturers, the software reduces its adaptability to multi-vendor systems. To overcome these limitations, a vendor-neutral interface known as the Standard Robot Command Interface (SRCI) was introduced, offering a client-server architecture based on Profinet [20]. The SRCI is also limited to the fieldbus Profinet and only the PLC manufacturer Siemens supports the logic on the PLC. Despite SRCI being a universal framework, it does not support advanced robotic motions like spline-based and force-controlled motions. Therefore, the SRCI is unsuitable for skill-based software with advanced robotic motions required for performing precise or complex tasks [21].

## 3 Need for Action

The current robotic skill implementations are largely limited to basic linear (LIN) and point-to-point (PTP) motions, which do not fully meet the demands of complex industrial tasks, like welding, gluing or sensitive assembly. Therefore, it is essential to develop standardized skills for all the advanced robotic motions provided by the robot manufacturers such as circular motion, spline motion and force-based motion. Further the existing robotic interfaces are either specific to a vendor or limited in advanced robotic motions functionalities like spline-based or force-based. This work addresses these limitations through an efficient dual-layered skill programming approach, ensuring efficient, flexible and scalable robotic skills.

To ensure standardization of the interface, every robotic skill must possess common motion parameters like final position in cartesian coordinate frame, velocity, acceleration, tool frame and base frame. Further, the individual skill like *MoveCircularSkill* will require an additional parameter for intermediate point on the circular path, which most industrial robot manufacturers support, e. g. ABB, KUKA, Universal Robots. Similarly, the additional function to the skill can be added to perform the skill motion with respect to joint space. Therefore, providing this additional skill function will enable the robot to achieve the desired pose.

A new *MoveContinuousSkill* is needed to meet the requirements described above for a manufacturer-independent and high-performance robot skill control system consisting of various motion primitives. This skill is particularly beneficial in applications where the robot passes through multiple waypoints without stopping, such as material handling, welding and sorting or where the operation requires precise control at the final position. Unlike conventional skills, this approach enables approximation of intermediate points to create smoother, quicker and energy-efficient motion. To support this functionality, the *MoveContinuousSkill* parameters must include required number of continuous motions, motion type and motion parameters for each motion and approximation distance for approximating the destination point of each motion. To achieve continuous motion, both a method for planning these motions and a way of controlling the robot with the planned motion sequences are required. Chapter 4 shows a methodical procedure for motion skill planning. This forms the basis for improving the speed of execution of skill-based robot programming. Chapter 5 shows the implementation of the interface for triggering a robot by PLC.

## 4 Method of Motion Skill Planning

The method for skill planning presented by [7] is expanded in this work to include new elements so that a continuous motion sequence with multiple motions for robots is possible. To plan the motion skills, the different requirements of the processes regarding path accuracy must be taken into account. For primary processes, such as welding, gluing or assembly processes, path accuracy is necessary to achieve the quality requirements and avoid collisions. For secondary motion sequences, a lower path accuracy is often sufficient. To optimize the planning of motion skills, different labeling of the skills is necessary. The following definition is used for this purpose:
- Blending: Smooth transition between different motion segments.
- Accurate path: The path is executed without blending.
- Accurate stop: Exact stop at the end-point of the path.

Figure 1 shows the planning method presented in [7] (gray background) with the extensions of the method from this work (white background).

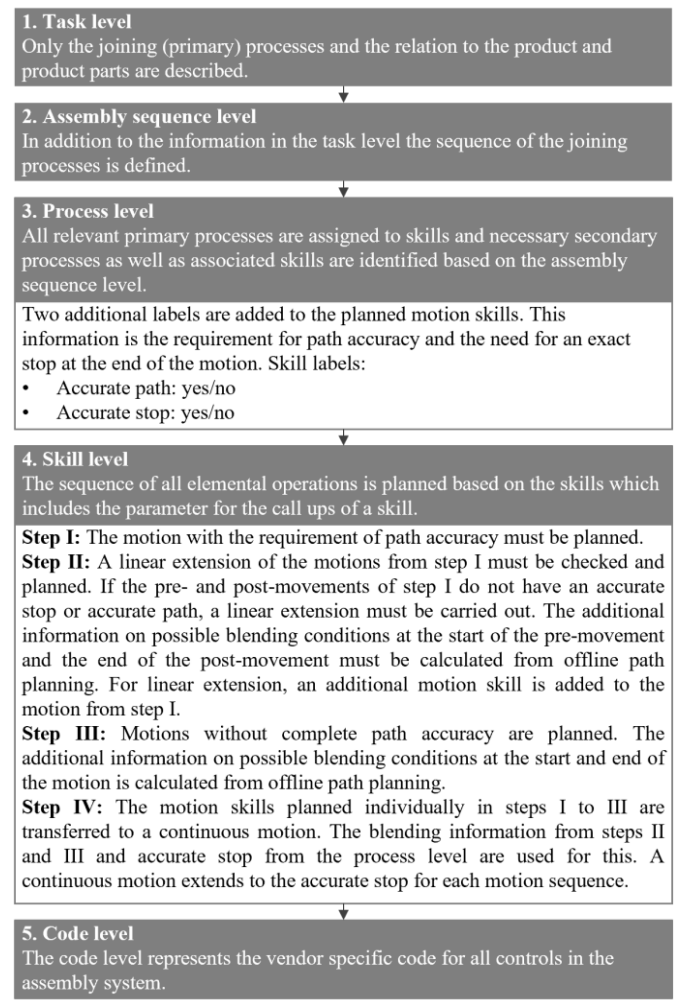

Figure 1: Extension of the skill planning method of [7] to include motion planning

## 5 Implementation of Motion Skill Execution

To execute the output of the motion skill planning method presented in chapter 4, an approach for controlling the robots is required. To ensure a high degree of industrial suitability, the control is carried out via a PLC.

### 5.1 *Dual-layered Architecture*

This work proposes a dual-layered architecture, where the primitive robot skill functions are programmed within the robot controller and the higher-level managerial robot skill functions are implemented in the PLC. Figure 2 shows a detailed overview of the *MoveContinuousSkill* implementation.

The first layer is the skill function in the PLC, which was implemented as a function block in IEC 61131-3. Each skill function block includes input ports for skill parameters as well as input and output mapping to the interface for communication with the robot controller. The function block also contains a skill state machine to control and monitor the skill execution, with each state defining the expected behavior. The skill parameters and commands are transmitted from the PLC to the robot controller through a fieldbus interface. Similarly, the feedback commands and current execution status are sent from the robot to the PLC via the same interface.

The second layer is the primitive robotic skill function, implemented directly in the robot controller. Here each skill has its own function called upon by the command from the corresponding skill in the PLC. The robot skill function assigns the received skill parameters to the position frame, velocity, acceleration, tool and base frame variables and executes the corresponding motion command. After the execution of the respective motion, the feedback is transmitted back to the PLC, ensuring real-time synchronization between both layers of the skill.

*5.2  MoveContinuousSkill Implementation*

The implementation of the new *MoveContinuousSkill* required additional logic in both layers of the skill execution framework. Since this skill is essential for executing sequential robot motions without interruptions, the robot controller must have future positions in advance to plan the upcoming motion trajectory. Every robot manufacturer provides this capability to perform this. For example, using KUKA robots this feature is known as Advance Pointer that is capable of pre-planning its trajectory based on upcoming positions [22]. Consequently, the *MoveContinuousSkill* must receive position data from the PLC in advance to ensure smooth motion execution.

The number of motions executed with the MoveContinuousSkill depends only on the PLC's memory space and can amount to several million. However, transmitting all motion parameters from the PLC to the robot in a single initial transfer as performed for other skills is not feasible due to the fieldbus interface input/output limitations of the robot controller. When using Profinet, a maximum of 256 bytes can be transferred. The data needed for one motion requires 44 bytes, so that a maximum count of five motions (*cntFC*) can be transmitted at once via the fieldbus. To overcome this, an update logic was integrated into the first level skill function. Initially, the first five motion parameters are sent to the robot. Once the skill execution begins, the update logic is triggered and the robot cyclically updates its execution index (*curExec*) after the completion of a motion. When the first motion is executed, the sixth position from the PLC is transferred to the first index of the motion queue of the robot using a First-In-First-Out (FIFO) update mechanism. This process continues cyclically until all motions (*motions*) in the

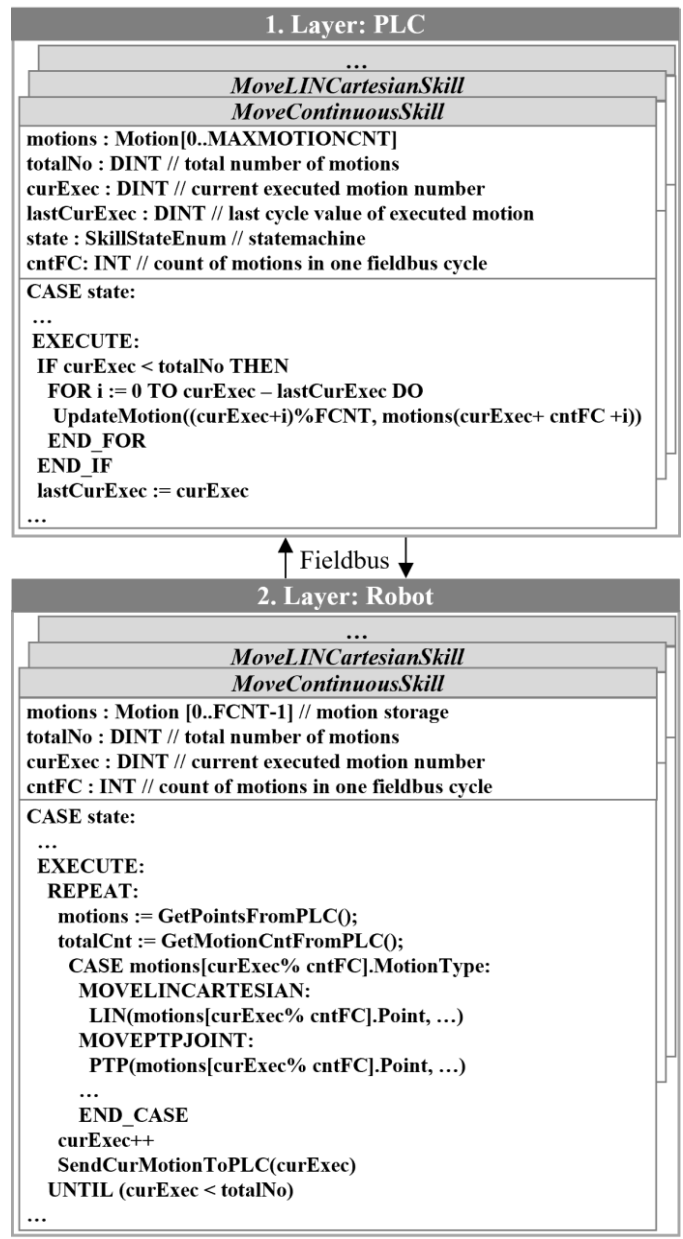

Figure 2: Code cutout from the dual-layered *MoveContinuousSkill*

*MoveContinuousSkill* have been executed. The total number of motions (*totalNo*) to be performed represents the end criterion of the motion sequence.

In the robotic skill function layer, the *MoveContinuousSkill* run a 'REPEAT UNTIL' loop to execute the required number of motions. Each motion has a specific index, enabling selection and execution of the desired motion type. This structured approach allows for seamless execution of various motion types with the desired motion parameters, ensuring precise and uninterrupted robotic motion. Various motion types have been implemented to meet the requirements of extensive robot programmability. These include *MoveLINCartesianSkill*, *MovePTPCartesianSkill*, *MovePTPJointSkill*, *MoveCircularSkill*. The robot controller of some manufacturers, e.g. KUKA, also allows the use of spline motions, which were implemented using a *MoveSplineSkill* motion type. Manufacturers of collaborative robots allow the use of force-based motions. These capabilities have been implemented as *MoveLINForceCartesianSkill*. This strategy eliminates the need for vendor-specific or vendor-neutral interfaces like mxAutomation, MotoLogix or SRCI, ensuring compatibility with diverse robotic systems. Furthermore, this approach is protocol-neutral, supporting communication protocols like Profinet, EtherNet/IP and others. By leveraging every motion capability of individual robots in their controllers and using the PLC for skill triggering, the proposed solution enhances flexibility and scalability, enabling seamless integration of heterogeneous robotic resources in FMS.

## 6 Application

An exemplary pick & place application was used to apply the planning method from chapter 4 and implement the dual-layered approach to robot control from chapter 5. Figure 3 shows the application example, which is similar to the pick & place application used by robot manufacturers [23, 24]. The robot picks a part from a 30 mm deep load carrier, moves the part to an assembly point 300 mm away and inserts the part. After opening the gripper, the robot returns to its starting position. To avoid a collision with the obstacle, the robot moves to a higher intermediate position.

### 6.1 Application of the Method for Motion Skill Planning

In this section, the planning method presented in figure 1 is applied, which consisting of the labelling at process level and the four steps at skill level.

An accurate stop must be performed at the end of the application to grip and release the gripper. The motion sequence consists of four accurate path motions. These are the motion of the gripper to the gripping position A, the removal of the component from the load carrier B, the insertion at the assembly point C and the removal of the gripper after the release of the gripper D. These four motions could be planned after determining the gripping point with step I. Motions A and C could be extended by a linear pre-movement in step II. Motions B and D could be extended in step II by a linear post movement. In step III, the secondary motions were planned as a PTP motion. In step IV, the individual motion skills of the pick & place application were transferred to three continuous motions. The first continuous motion runs from the starting point to the gripping position of the part. The second continuous motion begins after the part has been gripped up to the release position. The third continuous motion begins after the release and continues to the end point.

### 6.2 Application of the Motion Skill Execution

To demonstrate the transferability of the implementation of continuous motion skill execution presented in the chapter 5, it was implemented in various PLCs, fieldbuses and robots. Table 1 provides an overview of the used PLCs, fieldbuses and robots for two different experimental setups.

Table 1: Components used for the experimental setup

| Setup | PLC | Fieldbus | Robot |
|---|---|---|---|
| A | Siemens S7 1517 | Profinet | KUKA SCARA KR 6 R500 |
| B | B&R APC910 | EtherNet/IP | ABB IRB 1100 |

To evaluate the execution times of different programming methods, the sequence defined in figure 3 was programmed with three different execution types (ET):
1. Programmed in robot controller (RC), which allows comparison with the current procedure in the industry.
2. Programmed in PLC with single motions (SM), which leads to interrupts in the motion and allows comparison with the current state of research.
3. Programmed in PLC with the in chapter 5 presented *MoveContinuousSkill* (CM).

The update logic for continuous motion is only activated after five motions. The robot was not stopped at the gripping position and release position so that this was applied extensively and to reduce further influences on the measurement. The sequence was started by the PLC in all three execution types and the end was recognized on the PLC. The required sequence time was measured on the PLC. The cycle times of the PLC (1 ms), the fieldbus (1 ms) and the robot controller (4 ms) have only a minor influence on the execution times. As the same measurement method was used in all three execution types, possible deviations due to the cycle time have the same effect on all measurement series and are equalized by the number of repetitions. Table 2 shows the mean value of the average execution time (AET) and the mean absolute deviation (MAD) of the measurements performed. The measurement was repeated 25 times.

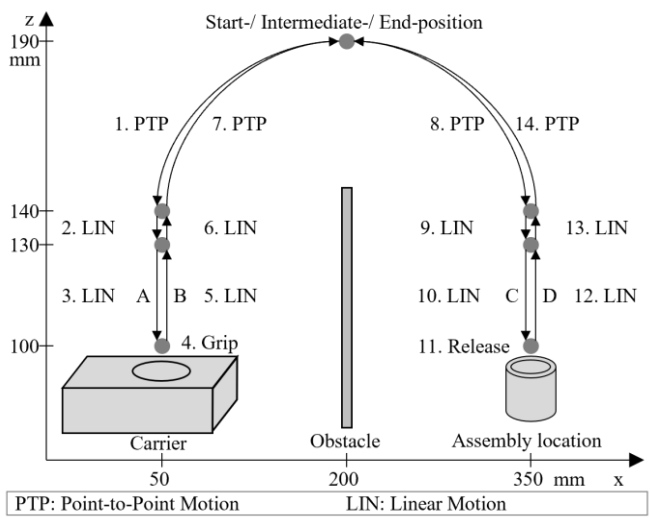

Figure 3: Pick & place application example

Table 2: Result of the measured execution times for the three execution types

| | Setup A | | Setup B | |
|---|---|---|---|---|
| ET | AET (ms) | MAD (ms) | AET (ms) | MAD (ms) |
| RC | 3725 | 3.6 | 7617 | 9.9 |
| SM | 6477 | 11.3 | 8971 | 20.8 |
| CM | 3729 | 3.3 | 7617 | 15.3 |

## 7  Discussion

The application of the planning method from chapter 4 has led to three continuous motion sequences of the individual motion segments and thus fulfilled the requirements for the planning method.

The measurement results of the different programming methods in table 2 demonstrate that the implementation of *MoveContinuousSkill* is very close to the execution times of direct programming in the robot controller. The *MoveContinuousSkill* is faster than the current approaches presented in the state of the research in skill-based programming, which is shown by the average execution time differences between execution type SM and CM in table 2. The percentage improvement of the AET (AET$_i$) can be calculated using formula (1) and corresponds to 42 % for setup A and 15 % for setup B. The difference in the AET between the two experimental setups shown in table 2 is due to the different kinematic architecture of the robots.

$$AET_i = \frac{SM - CM}{SM} \tag{1}$$

The overview of already realized implementations in different PLCs, fieldbuses and robots (shown in table 1) demonstrates the general transferability. However, it should also be noted that the programming on the robot side varies in complexity.

## 8  Summary and Outlook

In this work, the need for continuous motion of the robot for skill-based programming was identified. The existing approaches in the state of the art and research do not meet the requirements for speed, manufacturer independence and the use of possible motions. To meet these challenges, a method for planning the motion of skills was presented that enables the motion to be smoothed over. The implementation of an interface between PLC and robot was demonstrated and the update logic of the motions was shown. The measurements of the execution time for an exemplary pick & place motion demonstrate the speed improvement of skill-based programming compared to the state of the art and research.

Overall, this work demonstrates the methodical planning of motions and the control of these motions via this flexible skill architecture. The *MoveContinuousSkill* enables the elimination of stops in the robot motion, resulting in reduced energy consumption and minimized mechanical stress [19].
The motion planning method presented in this paper and the implemented interface between PLC and robot can be used to optimize trajectory planning and execution. This allows a finely planned trajectory to be transferred to the robot and executed. The software architecture for skill-based programming on the PLC will be presented in future work. Furthermore, an automated transfer of the control code to the PLC must take place for the application in the HMLV. In future work, the synchronous control of two different robot arms via the PLC is to be demonstrated.